\documentclass{article}
\usepackage{spconf,amsmath,graphicx, tikz, amsfonts}

\title{Sparse identification of nonlinear dynamics in the presence of library and system uncertainty}
\name{Andrew O'Brien}
\address{College of Computing \& Informatics, Drexel University }
\begin{document}
%
\maketitle
\begin{abstract}
The SINDy algorithm has been successfully used to identify the governing equations of dynamical systems from time series data. However, SINDy assumes the user has prior knowledge of the variables in the system and of a function library that can act as a basis for the system. In this paper, we demonstrate on real world data how the Augmented SINDy algorithm outperforms SINDy in the presence of system variable uncertainty. We then show SINDy can be further augmented to perform robustly when both kinds of uncertainty are present.

\end{abstract}
\begin{keywords}
Sparsity, causality, dynamical systems , SINDy
\end{keywords}
\section{Introduction}

The Sparse Identification of Nonlinear Dynamics (SINDy) algorithm, introduced by Brunton and colleagues \cite{Brunton:2016}, is designed to deduce the underlying equations of a dynamic system by analyzing data obtained from the system's temporal evolution. Widely adopted in engineering and signal analysis fields, this method has proven effective in deriving governing equations across various situations \cite{HorrocksBauch2020} \cite{Mangan:2016} \cite{Mukhachev2021}.

The algorithm requires a number of assumptions to hold which may not be realistic in real research scenarios. For example, a researcher attempting to use SINDy needs to know the exact variables in the system. If this assumption fails to hold and the researcher only knows a superset of the system variables, SINDy's performance can degrade considerably \cite{Obrien2023}. Previous research has show that SINDy can be augmented (Augmented SINDy), using techniques from causal discovery, to perform more robustly in the presence of this kind of uncertainty on synthetic data \cite{Obrien2023}. In this work, we show Augmented SINDy performs robustly on real world data. 

SINDy also assumes the researcher using it has prior knowledge of a function library which forms a basis of the governing equations of the system being studied. Past research has show that sparse coding can be adapted to learn missing functions of the SINDy dictionary \cite{OBrien2023BasisLearning}. However, to the best of the authors' knowledge, the question of how SINDy performs in the presence of both kinds of uncertainty remains unexplored in the literature. In this work, we demonstrate that SINDy's performance degrades significantly in the presence of both kinds of uncertainty, and show that if past augmentation methods for these kinds of uncertainty are combined, SINDy's performance becomes much more robust.

\section{Background}
\label{sec:background}

\subsection{Sparse Regression}
\label{ssec:sparse}

Regression, a key segment of machine learning, involves processing a data set $\{(X_{i}, Y_{i})\}_{i=1}^{N}$ along with a parameterized function $f(\cdot,\beta)$. The objective is to determine a parameter set $\beta^{*}$ so that $f(X_{i};\beta^{*}) \approx Y_{i}$. Typically, $X_{i},\beta \in \mathbb{R}^{k \times 1}$ and $Y_{i} \in \mathbb{R}$. A prevalent technique in regression is to assume that $f$ is linear and to identify the parameter set that minimizes the $l_{2}$ norm of the residual vector, representing the difference between the model predictions and the actual values. This methodology is encapsulated in Equation 1, where $X$ consists of the transposed $X_{i}$ data vectors and $Y$ encompasses the scalar values $Y_{i}$.

\begin{equation}
\beta^{*} = \underset{\beta}{\mathrm{argmin}} (\| X^{T}\beta - Y\|_{2})
\end{equation}

This approach, however, faces several challenges \cite{Brunton2022}. In scenarios where the regression is under-determined, Equation 1 may yield multiple valid solutions. In high-dimensional cases, interpreting $\beta$ becomes complex for stakeholders utilizing the model. Moreover, models characterized by an extensive number of parameters often struggle with generalization to new, unseen data. These issues can be addressed by adding an $l_{0}$ sparsity constraint to the optimization problem in Equation 1. However, direct implementation of this optimization is challenging, so the $l_{1}$ norm is often employed as a proxy for the $l_{0}$ norm, as illustrated in Equation 2.

\begin{equation}
\beta^{*} = \underset{\beta}{\mathrm{argmin}} (\|X\beta - Y\|_{2} + \lambda \|\beta\|_{1})
\end{equation}

\subsection{SINDy}

The Sparse Identification of Nonlinear Dynamics (SINDy) algorithm is designed to reconstruct dynamical systems from time series data \cite{Brunton:2016}. The formal input to SINDy is a data set $(\textbf{X}, \textbf{Y})$ representing samples from an autonomous dynamical system. The governing equations of the system are expressed as $\frac{d}{dt}\textbf{x}(t) = \textbf{f}(\textbf{x}(t); \boldsymbol \beta)$, where $t$ denotes time, $\textbf{x}(t)$ is the state of the system at time $t$, $\boldsymbol \beta$ symbolizes a set of system parameters, and $\textbf{f}$ is a function describing the system's state evolution at time $t$. Dynamical systems' governing functions are often linear combinations of a few simple functions \cite{Silva:2020}. SINDy approaches the task of learning these governing functions as a sparse regression problem.

More specifically, SINDy takes two matrices as input: $\textbf{X} = [x(t_1), ..., x(t_m)]^{T}$ and $\boldsymbol{ \dot{X} } = [\dot{x}(t_1), ..., \dot{x}(t_m)]^{T}$. Here, $x(t) \in \mathbb{R}^{n \times 1}$ represents the state of the system at time $t$, and $\dot{x}(t) \in \mathbb{R}^{n \times 1}$ is the measurement or estimation of the system's derivative at time $t$. From $\textbf{X}$, a library of candidate functions, denoted $\boldsymbol \theta (\textbf{X})$, is constructed. The optimization problem, shown in Equation 3, is then solved for each of the $N$ columns of $\boldsymbol{\dot{X}}$.

\begin{equation}
\xi_{k} = \underset{\xi_{k}^{'}}{\mathrm{argmin}} (\| \textbf{\.X}_{k} - \boldsymbol \theta (\textbf{X}) \xi_{k}^{'} \|_{2} + \lambda \| \xi_{k}^{'} \|_{1})
\end{equation}

Brunton et al. addressed the optimization problem in Equation 3 using the Sequential Thresholding Least Squares (STLS) method proposed in \cite{Brunton:2016}. STLS commences by executing a standard least squares regression of $\boldsymbol{\dot{X}_{k}}$ on $\boldsymbol{\theta (X)}$ to obtain preliminary values for $\xi_{k}$. A threshold $c$ is then selected, under which any element of $\xi_{k}$ smaller than $c$ is set to zero. Subsequently, a least squares regression is performed again on $\boldsymbol{\dot{X}_{k}}$
 against the non-zero coefficient columns of $\theta(\textbf{X})$. This process of regression and thresholding continues until the values of $\xi_{k}$ stabilize.

\subsection{SINDy Assumptions}

Efforts have been made to enhance the Sparse Identification of Nonlinear Dynamics (SINDy) algorithm by mitigating its reliance on certain limiting assumptions. Originally, SINDy presupposes that variables are measured in an appropriate coordinate system. To address this, Champion et al. modified the SINDy loss function by integrating an autoencoder, enabling the simultaneous discovery of both the correct coordinate system and the dynamical equations \cite{Champion2019DataDrivenDiscovery}. Recognizing SINDy's struggles with data corruption and outliers, Champion et al. devised a robust sparse optimization framework to better handle noisy data \cite{Champion2020UnifiedSparseOptimization}. To expand the variety of functions in the library, Kaheman et al. included rational functions \cite{Kaheman2020SINDYPI}. While SINDy initially focused on systems described by ordinary differential equations, Rudy et al. broadened its scope to encompass systems governed by partial differential equations \cite{Rudy2017DataDrivenDiscovery}.  Bakarji et al. enhanced SINDy for scenarios where some system variables are not measured \cite{Bakarji2023DiscoveringGoverningEquations}. O'Brien et al. augmented SINDy with techniques from causal discovery to make SINDy robust to the assumption that the research knows all the variables in a system as opposed to just a superset of them \cite{Obrien2023}. O'Brien et al. also showed that a researcher  could learn a basis for SINDy if one was not know a prior using a technique adapted from sparse coding \cite{OBrien2023BasisLearning}.

\section{Augmented SINDy \& Real World Data}

I conducted an evaluation of the Augmented SINDy algorithm's effectiveness using data from two real-world ecosystems. The initial case study focused on a predator-prey relationship involving the Canadian Lynx and the Snowshoe Hare populations \cite{odum2005fundamentals}. The second case involved analyzing a dataset from Sugihara et al. \cite{sugihara2012detecting}, which comprises data on Pacific sardine and northern anchovy landings. For both these ecosystems, additional variables were introduced to the analysis to simulate the type of uncertainty ecologists typically encounter in real-world data applications of SINDy.

Understanding and accurately modeling the interactions between species and their environments is crucial for ecologists and conservationists. Wildlife conservation initiatives often aim to implement measures that causally affect the population levels of certain species \cite{stokes2011introduction}. There's a growing interest in leveraging machine learning tools for gaining insights that could inform such conservation strategies \cite{tuia2022perspectives}. Although these methods have been beneficial, there remains significant scope for improvement, especially in applying causal machine learning techniques in these contexts. In this section, I demonstrate that Augmented SINDy is a promising tool for bridging this gap. Specifically, in Experiments 1 and 2, I test the hypothesis: ``Augmented SINDy surpasses traditional SINDy in learning causal relationships between variables in ecological systems."

\subsection{Experiments 1 \& 2}

In Experiment 1, I assessed the efficacy of Augmented SINDy over SINDy in discerning the causal connections within the Lynx-Hare ecosystem. The precise structural equations dictating the interactions between lynx and hare in this natural environment are unknown. Nonetheless, their established predator-prey dynamics imply some causal relationship. Additionally, Gaussian noise variables,  $x_{1}$ and  $x_{2}$, were incorporated, with $ x_{1} \sim \mathcal{N}(l_{mean}, l_{std}) )$ mirroring the distribution of l , and $x_{2} \sim \mathcal{N}(h_{mean}, h_{std})$ reflecting  h. The entire system for this experiment is represented as \( X = [l, h, x_1, x_2] \), simulating the challenge of analyzing population data from similar species in an environment with ambiguous causal links. An additional set of simulations with \( x_1, x_2 \sim \mathcal{N}(0, 1) \) was conducted to ensure the results were not unduly dependent on the synthetic variables' distribution.

The comparative analysis of SINDy and Augmented SINDy was based on the Fraction of Possible Incorrect Variables (FPIV) as an error metric. For example, since \( l \) is not causally influenced by \( x_1 \) and \( x_2 \), if \( l \)'s governing equation includes \( x_1 \) but not \( x_2 \), the FPIV would be 0.5. Including both \( x_1 \) and \( x_2 \) raises the FPIV to 1.0.

\begin{align}
    \dot{h} &= 0.1650h - 0.554l + 0.077x_1 + 0.134x_2 \\
    \dot{l} &= 0.137h - 0.114l - 0.038x_1 - 0.034x_2 \notag  \\
    \dot{x}_1 &= -0.022h - 0.040x_1 + 0.134x_2 \notag \\
    \dot{x}_2 &= -0.027h + 0.119l - 0.023x_1 \notag 
\end{align}

\begin{align}
    \dot{h} &= 0.186h - 0.426l \\
    \dot{l} &= 0.128h - 0.160l \notag    
\end{align}

This experiment was conducted over 10 iterations, each using different random samples from the Gaussian variables. In each iteration, five distinct lambda sparsity parameter values were tested, resulting in a total of fifty iterations. This process was repeated for both Gaussian variable distribution sets. The findings for SINDy and Augmented SINDy are documented in Tables 1 to 4, with example governing equations presented in Equations 4 and 5, As evidenced in Tables 1 through 4, Augmented SINDy markedly surpasses SINDy in FPIV performance and consistently recognizes the synthetic variables as non-causal.

\begin{table}[ht]
\centering
\resizebox{\columnwidth}{!}{
\begin{tabular}{| l | c | c | c | c | c |}
\hline
& $\lambda = .0900$ & $\lambda = .0810$ & $\lambda = .0729$ & $\lambda = .0656$  & $\lambda = .0590$ \\
\hline
Trial 1 & .083 & .000 & .000 & .333 & .083 \\
\hline
Trial2 & .083 & .083 & .083 & .083 & .167 \\
\hline
Trial3  & .333 & .000 & .167 & .083 & .250 \\
\hline
Trial 4 & .083 & .333 & .167 & .083 & .500 \\
\hline
Trial 5 & .250 & .333 & .083 & .333 & .417 \\
\hline
Trial 6 & .083 & .250 & .250 & .083 & .083 \\
\hline
Trial 7 & .083 & .083 & .167 & .333 & .083 \\
\hline
Trial 8 & .083 & .250 & .167 & .333 & .417 \\
\hline
Trial 9 & .250 & .250 & .000 & .333 & .333 \\
\hline
Trial 10 & .000 & .083 & .333 & .167 & .250 \\
\hline
Average & $.133 \pm (.105)$  & $.167 \pm (.130)$  & $.142 \pm (.104)$ & $.216 \pm(.125)$ & $.258 \pm(.154)$ \\
\hline
\end{tabular}
}
\caption{FPIV values for SINDy on the lynx and hare system with $x_{1} \sim \mathcal{N}(l_{\text{mean}}, l_{\text{std}})$ and $x_{2} \sim \mathcal{N}(h_{\text{mean}}, h_{\text{std}}).$}

\end{table}

\begin{table}[ht]
\centering
\resizebox{\columnwidth}{!}{
\begin{tabular}{| l | c | c | c | c | c |}
\hline
& $\lambda = .0900$ & $\lambda = .0810$ & $\lambda = .0729$ & $\lambda = .0656$  & $\lambda = .0590$ \\
\hline
Trial 1 & .000 & .000 & .000 & .000 & .000 \\
\hline
Trial 2 & .000 & .000 & .000 & .000 & .000 \\
\hline
Trial 3 & .000 & .000 & .000 & .000 & .000 \\
\hline
Trial 4 & .000 & .000 & .000 & .000 & .000 \\
\hline
Trial 5 & .000 & .000 & .000 & .000 & .000 \\
\hline
Trial 6 & .000 & .000 & .000 & .000 & .000 \\
\hline
Trial 7 & .000 & .000 & .000 & .000 & .000 \\
\hline
Trial 8 & .000 & .000 & .000 & .000 & .000 \\
\hline
Trial 9 & .000 & .000 & .000 & .000 & .000 \\
\hline
Trial 10 & .000 & .000 & .000 & .000 & .000 \\
\hline
Average & $.000 \pm (.000)$  & $.000 \pm (.000)$  & $.000 \pm (.000)$ & $.000 \pm(.000)$ & $.000 \pm(.000)$ \\
\hline
\end{tabular}
}
\caption{FPIV values for Augmented SINDy on the lynx and hare system with $x_{1} \sim \mathcal{N}(l_{mean}, l_{std})$ and $x_{2} \sim \mathcal{N}(h_{mean}, h_{std})$.}

\end{table}

\begin{table}[ht]
\centering
\resizebox{\columnwidth}{!}{
\begin{tabular}{| l | c | c | c | c | c |}
\hline
& $\lambda = .0900$ & $\lambda = .0810$ & $\lambda = .0729$ & $\lambda = .0656$  & $\lambda = .0590$ \\
\hline
Trial 1 & .500 & .417 & .500 & .333 & .500 \\
\hline
Trial2 & .417 & .333 & .500 & .667 & .583 \\
\hline
Trial3  & .500 & .667 & .333 & .583 & .583 \\
\hline
Trial 4 & .500 & .500 & .667 & .583 & .500 \\
\hline
Trial 5 & .583 & .500 & .333 & .667 & .500 \\
\hline
Trial 6 & .417 & .417 & .667 & .667 & .333 \\
\hline
Trial 7 & .5 & .583 & .417 & .583 & .5 \\
\hline
Trial 8 & .583 & .333 & .667 & .583 & .417 \\
\hline
Trial 9 & .333 & .667 & .417 & .500 & .417 \\
\hline
Trial 10 & .500 & .583 & .500 & .500 & .583 \\
\hline
Average & $.133 \pm (.105)$  & $.167 \pm (.130)$  & $.142 \pm (.104)$ & $.216 \pm(.125)$ & $.258 \pm(.154)$ \\
\hline
\end{tabular}
}
\caption{FPIV values for SINDy on the lynx and hare system with $x_{1} \sim \mathcal{N}(0, 1)$ and $x_{2} \sim \mathcal{N}(0, 1)$.}

\end{table}

\begin{table}[ht]
\centering
\resizebox{\columnwidth}{!}{
\begin{tabular}{| l | c | c | c | c | c |}
\hline
& $\lambda = .0900$ & $\lambda = .0810$ & $\lambda = .0729$ & $\lambda = .0656$  & $\lambda = .0590$ \\
\hline
Trial 1 & .000 & .000 & .000 & .000 & .000 \\
\hline
Trial2 & .000 & .000 & .000 & .000 & .000 \\
\hline
Trial3  & .000 & .000 & .000 & .000 & .000 \\
\hline
Trial 4 & .000 & .000 & .000 & .000 & .000 \\
\hline
Trial 5 & .000 & .000 & .000 & .000 & .000 \\
\hline
Trial 6 & .000 & .000 & .000 & .000 & .000 \\
\hline
Trial 7 & .000 & .000 & .000 & .000 & .000 \\
\hline
Trial 8 & .000 & .000 & .000 & .000 & .000 \\
\hline
Trial 9 & .000 & .000 & .000 & .000 & .000 \\
\hline
Trial 10 & 000 & .000 & .000 & .000 & .000 \\
\hline
Average & $.000 \pm (.000)$  & $.000 \pm (.000)$  & $.000 \pm (.000)$ & $.000 \pm(.000)$ & $.000 \pm(.000)$ \\
\hline
\end{tabular}
}
\caption{FPIV values for Augmented SINDy on the lynx and hare system with $x_{1} \sim \mathcal{N}(0, 1)$ and $x_{2} \sim \mathcal{N}(0, 1)$.}

\end{table}

For the subsequent real-world example, I employed a dataset from Sugihara et al. \cite{sugihara2012detecting} regarding pacific sardine and northern anchovy landings. The causal interaction between these species is contested \cite{murphy1964species}, with some attributing their population correlation to a third variable \cite{lasker1983proceedings}. Research indicates a lack of causal interaction \cite{jacobson1995stock}. Therefore, an effective system identification algorithm should infer governing equations that show no interaction between these species.

Experiment 2 was conducted in a manner similar to Experiment 1, utilizing species data for sardines and anchovies. The findings are presented in Table 5 through Table 8. Examples of the governing equations deduced by SINDy and Augmented SINDy are shown in Equations 6 and 7. Similar to the lynx and hare dataset, Augmented SINDy significantly surpasses SINDy in performance. This provides strong evidence that in scenarios where researchers only know a superset of variables within the system, regular SINDy proves inadequate, whereas Augmented SINDy delivers dependable results.

\begin{align}
\dot{a} &= 0.126x_2 \\
\dot{s} &= -0.707as + 0.162sx_1 + 0.249x_1x_2 - 0.143s^2 \notag \\
\dot{x}_1 &= 0.145s - 0.355as - 0.197sx_1 - 0.118x_1x_2 \notag \\
\dot{x}_2 &= -0.110s + 0.275as + 0.167ax_2 + 0.200sx_1 + 0.166x_1x_2 \notag
\end{align}

\begin{align}
\dot{a} &= 0.126x_2 
\end{align}

\begin{table}[ht]
\centering
\resizebox{\columnwidth}{!}{
\begin{tabular}{| l | c | c | c | c | c |}
\hline
& $\lambda = .0900$ & $\lambda = .0810$ & $\lambda = .0729$ & $\lambda = .0656$  & $\lambda = .0590$ \\
\hline
Trial 1 & .923 & .714 & .929 & .929 & .786 \\
\hline
Trial2 & .643 & .786 & .643 & 1.00 & 1.00 \\
\hline
Trial3  & .786 & .857 & .929 & .786 & .929 \\
\hline
Trial 4 & .429 & .714 & .857 & .857 & .857 \\
\hline
Trial 5 & .714  & .714 & .643 & .929 & 1.00 \\
\hline
Trial 6 & .786 & .786 & .929 & .929 & 1.00 \\
\hline
Trial 7 & .857 & .643 & .643 & .714 & 1.00 \\
\hline
Trial 8 & 1.00 & .714 & .786 & .857 & .786 \\
\hline
Trial 9 & .786 & .714 & .786 & 1.00 & .857\\
\hline
Trial 10 & .786 & .857 & .857 & .714 & .929 \\
\hline
Average & $.077 \pm (.157)$  & $.750 \pm (.069)$  & $.802 \pm (.123)$ & $.872 \pm(.106)$ & $.914 \pm(.088)$ \\
\hline
\end{tabular}
}
\caption{FPIV values for SINDy on the sardine and anchovie system with $x_{1} \sim \mathcal{N}(a_{mean}, a_{std})$ and $x_{2} \sim \mathcal{N}(s_{mean}, s_{std})$.}

\end{table}

\begin{table}[ht]
\centering
\resizebox{\columnwidth}{!}{
\begin{tabular}{| l | c | c | c | c | c |}
\hline
& $\lambda = .0900$ & $\lambda = .0810$ & $\lambda = .0729$ & $\lambda = .0656$  & $\lambda = .0590$ \\
\hline
Trial 1 & .000 & .000 & .000 & .000 & .000 \\
\hline
Trial2 & .000 & .000 & .000 & .000 & .000 \\
\hline
Trial3  & .000 & .000 & .000 & .000 & .000 \\
\hline
Trial 4 & .000 & .000 & .000 & .000 & .000 \\
\hline
Trial 5 & .000  & .000 & .000 & .000 & .000 \\
\hline
Trial 6 & .000 & .000 & .000 & .000 & .000 \\
\hline
Trial 7 & .000 & .000 & .000 & .000 & .000 \\
\hline
Trial 8 &  .000 & .000 & .000 & .000 & .000 \\
\hline
Trial 9 & .000 & .000 & .000 & .000 & .000\\
\hline
Trial 10 & .000 & .000 & .000 & .000 & .000 \\
\hline
Average & $.000 \pm (.000)$  & $.000 \pm (.000)$  & $.000 \pm (.000)$ & $.000 \pm(.000)$ & $.000 \pm(.000)$ \\
\hline
\end{tabular}
}
\caption{FPIV values for Augmented SINDy on the sardine and anchovie system with $x_{1} \sim \mathcal{N}(a_{mean}, a_{std})$ and $x_{2} \sim \mathcal{N}(s_{mean}, s_{std})$.}

\end{table}

\begin{table}[ht]
\centering
\resizebox{\columnwidth}{!}{
\begin{tabular}{| l | c | c | c | c | c |}
\hline
& $\lambda = .0900$ & $\lambda = .0810$ & $\lambda = .0729$ & $\lambda = .0656$  & $\lambda = .0590$ \\
\hline
Trial 1 & .714 & .786 & .929 & .714 & .857 \\
\hline
Trial2 & .500 & .929 & .786 & .786 & .929 \\
\hline
Trial3  & .714 & .714 & .786 & .786 & .786 \\
\hline
Trial 4 & .714 & .857 & .786 & .929 & 1.00 \\
\hline
Trial 5 & .857 & .857 & 1.00 & 1.00 & .786 \\
\hline
Trial 6 & .714 & .571 & .857 & .857 & .857 \\
\hline
Trial 7 & .786 & .500 & .857 & .857 & .929 \\
\hline
Trial 8 & .714 & .786 & .857 & .857 & .929 \\
\hline
Trial 9 & .714 & .857 & .857 & 1.00 & .929 \\
\hline
Trial 10 & .500 & .786 & .786 & .786 & .857 \\
\hline
Average & $.693 \pm (.112)$  & $.764 \pm (.135)$  & $.850 \pm (.071)$ & $.857 \pm(.095)$ & $.886 \pm(.069)$ \\
\hline
\end{tabular}
}
\caption{FPIV values for SINDy on the sardine and anchovie system with $x_{1} \sim \mathcal{N}(0, 1)$ and $x_{2} \sim \mathcal{N}(0, 1)$.}

\end{table}

\begin{table}[ht]
\centering
\resizebox{\columnwidth}{!}{
\begin{tabular}{| l | c | c | c | c | c |}
\hline
& $\lambda = .0900$ & $\lambda = .0810$ & $\lambda = .0729$ & $\lambda = .0656$  & $\lambda = .0590$ \\
\hline
Trial 1 & .000 & .000 & .000 & .000 & .000 \\
\hline
Trial2 & .000 & .000 & .000 & .000 & .000 \\
\hline
Trial3  & .000 & .000 & .000 & .000 & .000 \\
\hline
Trial 4 & .000 & .000 & .000 & .000 & .000 \\
\hline
Trial 5 & .000 & .000 & .000 & .000 & .000 \\
\hline
Trial 6 & .000 & .000 & .000 & .000 & .000 \\
\hline
Trial 7 & .000 & .000 & .000 & .000 & .000 \\
\hline
Trial 8 & .000 & .000 & .000 & .000 & .000 \\
\hline
Trial 9 & .000 & .000 & .000 & .000 & .000 \\
\hline
Trial 10 & .000 & .000 & .000 & .000 & .000 \\
\hline
Average & $.000 \pm (.000)$  & $.000 \pm (.000)$  & $.000 \pm (.000)$ & $.000 \pm(.000)$ & $.000 \pm(.000)$ \\
\hline
\end{tabular}
}
\caption{FPIV values for Augmented SINDy on the sardine and anchovie system with $x_{1} \sim \mathcal{N}(0, 1)$ and $x_{2} \sim \mathcal{N}(0, 1)$.}

\end{table}

\section{System \& Library Uncertainty}
\label{sec:print}
\subsection{Experiment 3}

The subsequent inquiry focused on the performance of Augmented SINDy when applied to a dictionary that is partially inferred, as referenced in \cite{OBrien2023BasisLearning}. This scenario aligns with a relaxation of both primary assumptions, leading to uncertainties in both the library of functions and the system variables. I generated data using the Lorenz \cite{Lorenz:1963}, Mankiw-Romer-Weil \cite{Mankiw:1992}, Fitzhugh-Nagumo \cite{Izhikevich:2010}, Pendulum \cite{Strogatz:2015}, and SIR models \cite{Thurner:2018}, incorporating noise variables in some instances. For each model, I consider an unknown subset of the dictionary, $\{f_1 (X_1\subseteq X),\ldots,f_k (X_k\subseteq X)\}\subseteq\theta(X)$. An auxiliary basis, $N=\{N_{k+1} (X_{k+1}\subseteq X),\ldots,N_L (X_L\subseteq X)\}$, was learned through sparse coding. The known part of the function dictionary, $F$, was then augmented with $N$ to form a new dictionary, $\hat{\theta}=F\cup N$. Utilizing $\hat{\theta}\cup Z$ as its dictionary, where $Z$ represents noise variables, the Augmented SINDy algorithm was employed. For comparative purposes, the SINDy algorithm was also executed using $F\cup Z$, simulating standard SINDy's application in contexts of dual uncertainty.

As a metric for performance, I used the proportion of accurately selected dictionary elements. Let's define a system's dictionary as $\theta^* = \theta \cup Z$. The system's $k$ equations can be represented by $k$ binary vectors, denoted as $\Xi = [\xi_1, \ldots, \xi_k]$. Correspondingly, the $k$ vectors deduced from either SINDy or Augmented SINDy are represented as $\Xi' = [\xi_1', \ldots, \xi_k']$. The Fraction of Dictionary Elements Correctly Selected (FDECS) is then defined as seen in Equations 8 and 9. The underlying hypothesis tested is: \textbf{Augmented SINDy, utilizing F $\cup$ N as a dictionary, achieves higher FDECS scores compared to SINDy, which employs F as its dictionary}.

\begin{equation}
FDES = \frac{\sum_{i=1}^k \|\xi_i - \text{SELECT}(\xi_i')\|_1}{k}
\end{equation}

\begin{equation}
\text{SELECT}(v) = v', \quad v_i' = 
\begin{cases}
    1, & \text{if } v_i \neq 0 \\
    0, & \text{if } v_i = 0
\end{cases}
\end{equation}

The performance of Augmented SINDy and regular SINDy with respect to FDES are given in Table 9 through Table 13. 

\begin{table}[h!]
\centering
\begin{tabular}{|c|c|c|}
\hline
\textbf{$|N|$} & \textbf{SINDy} & \textbf{Augmented SINDy}\\
\hline
1 & .60 & 1.00 \\
\hline
2 & .54 & .88 \\
\hline
3 & .58 & .92 \\
\hline
4 & .65 & .88 \\
\hline
5 & .63 & .98 \\
\hline
\end{tabular}
\caption{FDES values for the Lorenz system.}

\end{table}

\begin{table}[h!]
\centering
\begin{tabular}{|c|c|c|}
\hline
\textbf{$|N|$} & \textbf{SINDy} & \textbf{Augmented SINDy}\\
\hline
1 & .15 & .90 \\
\hline
2 & .50 & .90 \\
\hline
3 & .55 & .85 \\

\hline
\end{tabular}
\caption{FDES values for the MRW system.}

\end{table}

\begin{table}[h!]
\centering
\begin{tabular}{|c|c|c|}
\hline
\textbf{$|N|$} & \textbf{SINDy} & \textbf{Augmented SINDy}\\
\hline
1 & .68 & .89 \\
\hline
2 & .71 & .85 \\
\hline
3 & .61 & .96 \\
\hline
4 & .54 & .89 \\
\hline
5 & .61 & .75 \\
\hline
\end{tabular}
\caption{FDES values for the FN system.}

\end{table}

\begin{table}[h!]
\centering
\begin{tabular}{|c|c|c|}
\hline
\textbf{$|N|$} & \textbf{SINDy} & \textbf{Augmented SINDy}\\
\hline
1 & .55 & .90 \\
\hline
2 & .55 & .90 \\
\hline

\end{tabular}
\caption{FDES values for the SIR system.}
\end{table}

\begin{table}[h!]
\centering
\begin{tabular}{|c|c|c|}
\hline
\textbf{$|N|$} & \textbf{SINDy} & \textbf{Augmented SINDy}\\
\hline
1 & .55 & .90 \\
\hline
2 & .55 & .90 \\
\hline

\end{tabular}
\caption{FDES values for the Pendulum system.}

\end{table}

\section{Conclusion}
\label{sec:page}

The performance of the Augmented SINDy algorithm, when using the expanded dictionary obtained through sparse coding, was consistently superior to that of the standard SINDy in all tested systems. This indicates that the combined use of the Augmented SINDy algorithm and the basis learning algorithm can significantly enhance the efficacy of SINDy, especially in situations where there is uncertainty regarding both the dictionary and the system variables. Augmented SINDy also outperforms on real world data. These results should provide strong evidence that extensions of SINDy can be used even in situations of realistic uncertainty.

\bibliographystyle{IEEEbib}
\bibliography{refs}

\begin{thebibliography}{10}

\bibitem{Brunton:2016}
Steven~L. Brunton, Joshua~L. Procter, and J.~Nathan Kutz,
\newblock ``Discovering governing equations from data by sparse identification of nonlinear dynamical systems,''
\newblock {\em PNAS}, vol. 113, no. 15, pp. 3923--3937, 2016.

\bibitem{HorrocksBauch2020}
J.~Horrocks and C.~T. Bauch,
\newblock ``Algorithmic discovery of dynamic models from infectious disease data,''
\newblock {\em Scientific Reports}, vol. 10, 2020.

\bibitem{Mangan:2016}
Niall~M. Mangan, Steven~L. Brunton, Joshua~L. Proctor, and J.~Nathan Kutz,
\newblock ``Inferring biological networks by sparse identification of nonlinear dynamics,''
\newblock {\em IEEE Transactions on Molecular, Biological and Multi-Scale Communications}, vol. 2, no. 1, pp. 169259--169271, 52-63.

\bibitem{Mukhachev2021}
P.~Mukhachev, Z.~Sukhov, T.~Sadretdinov, and A.~Ivanov,
\newblock ``Evaluation of ml algorithms for system dynamics identification of aircraft pressure control system,''
\newblock {\em PHMEC ON F}, vol. 6, no. 1, pp. 7, Jun 2021.

\bibitem{Obrien2023}
Andrew O’Brien, Rosina Weber, and Edward Kim,
\newblock ``Investigating sindy as a tool for causal discovery in time series signals,''
\newblock in {\em ICASSP 2023 - 2023 IEEE International Conference on Acoustics, Speech and Signal Processing (ICASSP)}, 2023, pp. 1--5.

\bibitem{OBrien2023BasisLearning}
A.~O'Brien, R.~Weber, and E.~Kim,
\newblock ``Basis learning for dynamical systems in the presence of incomplete scientific knowledge,''
\newblock in {\em 2023 IEEE Sixth International Conference on Artificial Intelligence and Knowledge Engineering (AIKE)}, Laguna Hills, CA, USA, 2023.

\bibitem{Brunton2022}
Steven~L. Brunton and J.~Nathan Kutz,
\newblock {\em Data-Driven Science and Engineering: Machine Learning, Dynamical Systems, and Control},
\newblock Cambridge University Press, 2 edition, 2022.

\bibitem{Silva:2020}
Brian~M. de~Silva, David~M. Higdon, Steven~L. Brunton, and J.~Nathan Kutz,
\newblock ``Discovering of physics from data: Universal laws and discrepancies,''
\newblock {\em Frontiers in Artificial Intelligence}, vol. 3, 2020.

\bibitem{Champion2019DataDrivenDiscovery}
Katie Champion, Bethany Lusch, J.~Nathan Kutz, and Steven~L. Brunton,
\newblock ``Data-driven discovery of coordinates and governing equations,''
\newblock {\em Proceedings of the National Academy of Sciences}, vol. 116, no. 45, pp. 22445--22451, 2019.

\bibitem{Champion2020UnifiedSparseOptimization}
K.~Champion, P.~Zheng, A.~Aravkin, S.~Brunton, and J.~Kutz,
\newblock ``A unified sparse optimization framework to learn parsimonious physics-informed models from data,''
\newblock {\em IEEE Access}, vol. 8, pp. 169259--169271, Jan 2020.

\bibitem{Kaheman2020SINDYPI}
K.~Kaheman, J.~Nathan Kutz, and S.~L. Brunton,
\newblock ``Sindy-pi: A robust algorithm for parallel implicit sparse identification of nonlinear dynamics,''
\newblock {\em Proc. R. Soc. A}, vol. 476, pp. 20200279, 2020.

\bibitem{Rudy2017DataDrivenDiscovery}
Samuel~H Rudy, Steven~L Brunton, Joshua~L Proctor, and J~Nathan Kutz,
\newblock ``Data-driven discovery of partial differential equations,''
\newblock {\em Science Advances}, vol. 3, no. 4, pp. e1602614, 2017.

\bibitem{Bakarji2023DiscoveringGoverningEquations}
J.~Bakarji, K.~Champion, J.~Nathan Kutz, and S.~L. Brunton,
\newblock ``Discovering governing equations from partial measurements with deep delay autoencoders,''
\newblock {\em Proc. R. Soc. A}, vol. 479, pp. 20230422, 2023.

\bibitem{odum2005fundamentals}
E.~P. Odum and G.~W. Barrett,
\newblock {\em Fundamentals of Ecology},
\newblock Thomson Brooks/Cole, Belmont, CA, 5th edition, 2005.

\bibitem{sugihara2012detecting}
G.~Sugihara, R.~May, H.~Ye, C.~Hao Hsieh, E.~Deyle, M.~Fogarty, and S.~Munch,
\newblock ``Detecting causality in complex ecosystems,''
\newblock {\em Science}, vol. 338, no. 6106, pp. 496--500, 2012.

\bibitem{stokes2011introduction}
M.~Stokes,
\newblock ``An introduction to managing wild game populations,''
\newblock {\em Nature Education Knowledge}, vol. 2, no. 11, pp. 4, 2011.

\bibitem{tuia2022perspectives}
D.~Tuia, B.~Kellenberger, S.~Beery, et~al.,
\newblock ``Perspectives in machine learning for wildlife conservation,''
\newblock {\em Nature Communications}, vol. 13, pp. 792, 2022.

\bibitem{murphy1964species}
G.~I. Murphy and J.~D. Isaacs,
\newblock ``Species replacement in marine ecosystems with reference to the california current,''
\newblock {\em Minutes of Meeting Marine Research Committee}, vol. 7, no. 1, 1964.

\bibitem{lasker1983proceedings}
R.~Lasker and A.~MacCall,
\newblock ``Proceedings of the joint oceanographic assembly, halifax, august 1982: General symposia,''
\newblock in {\em Department of Fisheries and Oceans}, 1983, pp. 110--120.

\bibitem{jacobson1995stock}
Larry~D. Jacobson and Alec~D. MacCall,
\newblock ``Stock-recruitment models for pacific sardine ({Sardinops sagax}),''
\newblock {\em Canadian Journal of Fisheries and Aquatic Sciences}, vol. 52, no. 3, pp. 566--577, 1995.

\bibitem{Lorenz:1963}
Edward~Norton Lorenz,
\newblock ``Deterministic nonperiodic flow,''
\newblock {\em Journal of the Atmospheric Sciences}, vol. 20, no. 2, pp. 130--141, 1963.

\bibitem{Mankiw:1992}
N.~Gregory Mankiw, David Romer, and David~N. Weil,
\newblock ``A contribution to the empirics of economic growth,''
\newblock {\em The Quarterly Journal of Economics}, vol. 107, no. 2, pp. 407--437, 1992.

\bibitem{Izhikevich:2010}
Eugene~M. Izhikevich,
\newblock {\em Dynamical Systems in Neuroscience: The Geometry of Excitability and Bursting},
\newblock The MIT Press, 2010.

\bibitem{Strogatz:2015}
Steve~H. Strogatz,
\newblock {\em Nonlinear Dynamics and Chaos: With Applications to Physics, Biology, Chemistry, and Engineering},
\newblock CRC Press;, 2 edition, 2015.

\bibitem{Thurner:2018}
Stefan Thurner, Rudolf Hanel, and Peter Klimek,
\newblock {\em Introduction to the Theory of Complex Systems},
\newblock Oxford University Press;, 2018.

\end{thebibliography}

\end{document}